\documentclass[journal]{IEEEtranTIE}
\usepackage{graphicx}
\usepackage{cite}
\usepackage{picinpar}
\usepackage{amsmath}
\usepackage{url}
\usepackage{flushend}
\usepackage[latin1]{inputenc}
\usepackage{colortbl}
\usepackage{soul}
\usepackage{multirow}
\usepackage{pifont}
\usepackage{color}
\usepackage{alltt}
\usepackage[hidelinks]{hyperref}
\usepackage{enumerate}
\usepackage{siunitx}
\usepackage{breakurl}
\usepackage{epstopdf}
\usepackage{pbox}

\usepackage{adjustbox}
\usepackage{booktabs}
\usepackage{subfig}

\begin{document}
\title{	SpikingTac: A Miniaturized Neuromorphic Visuotactile Sensor for High-Precision Dynamic Tactile Imprint Tracking  }

\author{
	\vskip 1em
	
	Tianyu Jiang,
	Chaofan Zhang,
	Shaolin Zhang,
	Shaowei Cui*, \emph{Member, IEEE},
	\\ and Shuo Wang, \emph{Member, IEEE}

	\thanks{	
        *Corresponding Author.

		Tianyu Jiang and Chaofan Zhang are with the Institute of Automation, Chinese Academy of Sciences, Beijing 100190, China, and also with the School of Artificial Intelligence, University of Chinese Academy of Sciences, Beijing 100049,
China (e-mail: jiangtianyu2023@ia.ac.cn; chaofan.zhang@ia.ac.cn).

		Shaolin Zhang and Shaowei Cui are with the State Key Laboratory of Multimodal Artificial Intelligence Systems, Institute of Automation, Chinese Academy of Sciences, Beijing 100190, China (e-mail: zhangshaolin2015@ia.ac.cn; shaowei.cui@ia.ac.cn).
		
		Shuo Wang is with the State Key Laboratory of Multimodal Artificial Intelligence Systems, Institute of Automation, Chinese Academy of Sciences, Beijing 100190, China, also with the School of Artificial Intelligence, University of Chinese Academy of Sciences, Beijing 100049, China, and also with the Center for Excellence in Brain Science and Intelligence Technology, Chinese Academy of Sciences, Shanghai 200031, China (e-mail: shuo.wang@ia.ac.cn).
	}
}

\maketitle

\begin{abstract}
High-speed event-driven tactile sensors are essential for achieving human-like dynamic manipulation, yet their integration is often limited by the bulkiness of standard event cameras. This paper presents SpikingTac, a miniaturized, highly integrated neuromorphic tactile sensor featuring a custom standalone event camera module, achieved with a total material cost of less than \$150. We construct a global dynamic state map coupled with an unsupervised denoising network to enable precise tracking at a 1000~Hz perception rate and 350~Hz tracking frequency. Addressing the viscoelastic hysteresis of silicone elastomers, we propose a hysteresis-aware incremental update law with a spatial gain damping mechanism. Experimental results demonstrate exceptional zero-point stability, achieving a 100\% return-to-origin success rate with a minimal mean bias of 0.8039 pixels, even under extreme torsional deformations. In dynamic tasks, SpikingTac limits the obstacle-avoidance overshoot to 6.2~mm, representing a 5-fold performance improvement over conventional frame-based sensors. Furthermore, the sensor achieves sub-millimeter geometric accuracy, with Root Mean Square Error (RMSE) of 0.0952~mm in localization and 0.0452~mm in radius measurement.
\end{abstract}

\begin{IEEEkeywords}
Event-based sensing, neuromorphic tactile sensor, perception for collision detection, sub-millimeter geometric perception, unsupervised denoising.
\end{IEEEkeywords}

{}

\definecolor{limegreen}{rgb}{0.2, 0.8, 0.2}
\definecolor{forestgreen}{rgb}{0.13, 0.55, 0.13}
\definecolor{greenhtml}{rgb}{0.0, 0.5, 0.0}

\section{Introduction}


\IEEEPARstart{T}{actile} sensing is an essential modality for both humans and robots to perceive and interact with the physical environment \cite{gelsight}. Recently, neuromorphic vision-based tactile sensors have gained significant traction \cite{review}, driven by the rapid advancement of event cameras. Unlike conventional frame-based sensors, event cameras provide microsecond-level temporal resolution, high dynamic range (HDR), and minimal data redundancy \cite{review1, survey}. These capabilities have enabled robust performance across diverse robotic tasks, including object classification \cite{object recognition}, gesture recognition \cite{gesture recognition}, and texture detection \cite{texture recognition}.


However, the widespread integration of neuromorphic tactile sensors into dexterous robotic hands remains hindered by the bulky form factors and prohibitive costs of commercial event camera modules \cite{VBTS_review}. Although prototypes like Evatac \cite{evetac}, E-BTS \cite{ebts}, and NeuTac \cite{neutac} demonstrate the potential of event-based sensing, their physical dimensions limit practical deployment in space-constrained applications. Inspired by the open-source methodology of 9DTact \cite{9dtact}, we develop SpikingTac: a compact, low-cost neuromorphic tactile sensor. By integrating a miniaturized event camera with a custom driver board, our design ensures robust optical isolation and stable internal illumination, achieving a high signal-to-noise ratio (SNR) for high-fidelity feature extraction.

\begin{figure}[!t]
 \centering
  \includegraphics[width=0.9\columnwidth]{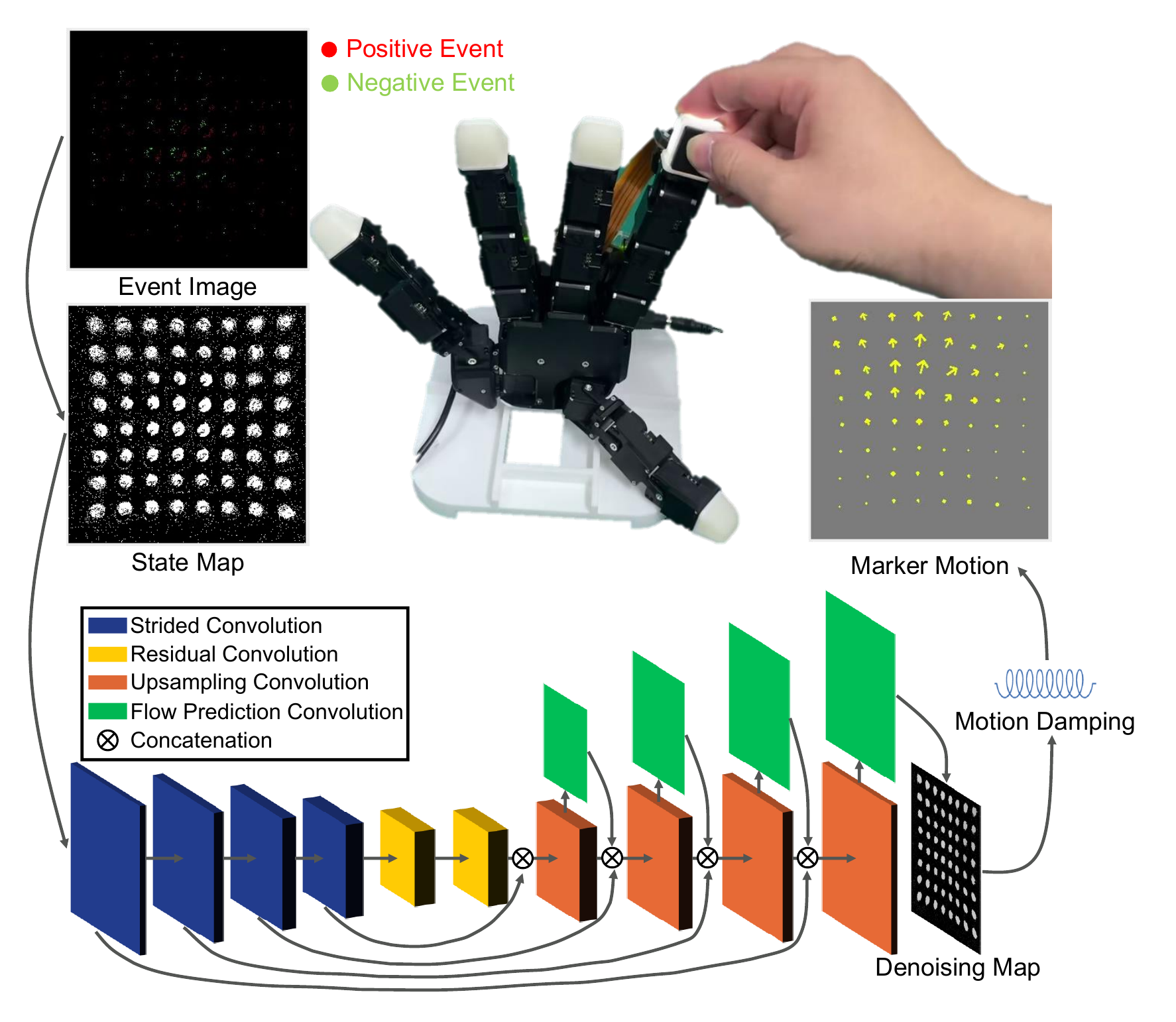}
 \caption{We present SpikingTac, a miniaturized neuromorphic visuotactile tactile sensor designed for seamless integration into dexterous robotic hands. By leveraging dynamic state map reconstruction for tactile imprints, SpikingTac enables high-bandwidth dynamic perception and ensures robust marker tracking under rapid physical interactions.}
 \label{intro}
\end{figure}


A critical challenge in neuromorphic tactile sensing is the real-time tracking of markers embedded within the elastomeric gel \cite{visual tracking}. When external forces deform the sensor surface, marker displacements, captured as event streams, encode vital tactile information, including force vectors \cite{contact force measurements}, slip \cite{slip detection}, and contact geometry \cite{fast sensing}. However, since event cameras only respond to temporal intensity changes, retrieving absolute marker positions under quasi-static conditions is non-trivial \cite{sensitive}. Furthermore, marker-induced events are often corrupted by background noise \cite{noise}. To address these issues, we propose a global dynamic state map framework coupled with an unsupervised denoising network. This approach enables crisp extraction of marker patterns from sparse data, achieving a robust tracking frequency of 350 Hz.


Beyond tracking accuracy, the viscoelastic hysteresis of the silicone elastomer poses a significant challenge for precise dynamic measurements \cite{visco}. To mitigate this, we introduce a hysteresis-aware incremental update law based on a spatial gain decay mechanism. This law effectively attenuates residual offsets during the dynamic recovery phase, ensuring zero-point stability and enhancing sensor reliability during continuous physical interactions.


Furthermore, to fully exploit the native temporal resolution of neuromorphic vision, we implement a high-frequency perception framework. We develop a collision detection scheme that monitors statistical signatures in the raw event stream at 1000 Hz. By identifying step-changes in event counts within a 1-ms window, SpikingTac achieves reflex-like response, enabling robots to detect impacts with minimal, velocity-independent latency. This capability further facilitates precise active tactile exploration, resolving micro-scale geometric features-such as estimating the center and radius of industrial-grade holes with sub-millimeter accuracy-without requiring complex marker tracking.

The main contributions of this work are summarized as follows:

\begin{enumerate}[1)]
    \item  We design SpikingTac, a compact and highly integrated tactile sensor utilizing a custom standalone event camera module. This hardware optimization overcomes the bulkiness of commercial modules while maintaining robust optical isolation and high-SNR event output.
    
    \item  We propose a global dynamic state map coupled with an unsupervised denoising network for robust feature extraction at 350 Hz. To address the viscoelastic lag of silicone, a hysteresis-aware incremental update law with a spatial gain damping mechanism was introduced, achieving exceptional zero-point stability with a 100\% return-to-origin success rate.
    
    \item  We demonstrated a 1000 Hz perception rate that reduces post-impact overshoot to 6.2 mm--a 5-fold improvement over frame-based sensors. Additionally, the system enables high-precision tactile exploration, achieving sub-millimeter geometric accuracy in micro-scale localization and sizing.
\end{enumerate}

\section{Related Work}

\subsection{Vision-based Tactile Sensors: From Frames to Events}
Conventional vision-based tactile sensors, such as GelSight \cite{gelsight} and its derivatives \cite{gelsight1, gelsight2, gelsight3}, achieve high-resolution mapping through dense visual features. However, their performance in high-speed manipulation is often bottlenecked by low frame rates and motion blur \cite{bad rgb}. To address this, neuromorphic tactile sensing has emerged as a high-bandwidth alternative \cite{event_promise}. Recent developments include Evetac \cite{evetac}, which perceives vibrations up to 498 Hz with minimal data redundancy, and GelEvent \cite{gelevent}, which utilizes event-based thresholding for stable contact area estimation. Other designs have integrated PWM light sources to enhance marker tracking within event streams \cite{ebts}.

Despite these temporal gains, existing neuromorphic sensors remain constrained by bulky form factors and high hardware complexity, hindering their integration into multi-fingered robotic hands. Inspired by the open-source 9DTact \cite{9dtact}, this work proposes a miniaturized, cost-effective event-based sensor module optimized for seamless robotic deployment.

\subsection{Event-based Feature Tracking and Pattern Recognition}

Tracking patterns within the tactile elastomer is fundamental for downstream sensing tasks \cite{track_important}. Existing methodologies are primarily categorized into optimization-based and learning-based approaches. In the optimization domain, techniques include fitting time-varying circles via normal flow \cite{eKalibr}, and spatial matching between event streams and predefined models \cite{visual tracking}. Funk et al. further improved robustness by incorporating grid-based regularization \cite{evetac}. However, these methods often require sensitive parameter tuning and struggle with complex contact modes due to a lack of global contextual awareness.

Alternatively, learning-based approaches have gained traction. Ward-Cherrier et al. utilized Spiking Neural Networks for spatial encoding \cite{neurotac}, while Salah et al. employed continuous event time surfaces and unsupervised denoising for global pattern reconstruction \cite{neutac}. Despite promising qualitative results, the quantitative accuracy of marker tracking-particularly during high-speed, repetitive contact-remains under-explored. This work extends the unsupervised reconstruction paradigm, focusing on its generalizability and quantitative performance in high-dynamic scenarios.

\begin{table*}[!t]
	\renewcommand{\arraystretch}{1.3}
	\caption{Comparison of GelSight-Mini, GelStereo 2.0, GelEvent, Evetac, NeuTac, \cite{eroller} and SpikingTac. ($^*$Commodity price. $^\dagger$Estimated price.)}
	\centering
	\label{tab1}
	\begin{tabular*}{\textwidth}{@{\extracolsep{\fill}}ll ccccc}
		\hline\hline \\[-3mm]
		\textbf{Perception} & \textbf{Sensor} & \textbf{Dimension [$mm^3$]} $\downarrow$ & \textbf{Sensing Area [$mm^2$]} $\uparrow$ & \textbf{D/A Ratio} & \textbf{FPS} $\uparrow$ & \textbf{Cost [\$]} $\downarrow$ \\[1.6ex] \hline
		RGB-based & GelSight-Mini & $32 \times 28.5 \times 28 = 25536$ & $19 \times 15 = 285$ & 90 & 25 & 499$^*$ \\
		& GelStereo 2.0 & $30 \times 30 \times 29 = 26100$ & $23 \times 23 = 529$ & 49 & 32 & 114 \\ \hline
		Event-based & GelEvent & $40 \times 40 \times 96 = 153600$ & $32 \times 32 = 1024$ & 150 & 180 & 2275$^\dagger$ \\
		& Evetac & $32 \times 33 \times 65 = 68640$ & $19 \times 15 = 285$ & 241 & \textbf{1000} & 5000$^\dagger$ \\
		& NeuTac & $200 \times 120 \times 100 = 2400000$ & $2 \times \pi \times 10^2 = 628$ & 3820 & 300 & 8000$^\dagger$ \\
		& A. Khairi \cite{eroller} & $116 \times 148 \times 88 = 1510784$ & $30 \times 40 = 1200$ & 1259 & NA & 5000$^\dagger$ \\
		& \textbf{SpikingTac(Ours)} & \textbf{22 $\times$ 22 $\times$ 19 = 9196} & \textbf{20.6 $\times$ 20.6 = 424.36} & \textbf{22} & 350 & \textbf{150} \\ [1.4ex]
		\hline\hline
	\end{tabular*}
\end{table*}

\subsection{Viscoelasticity and Hysteresis Compensation}

The intrinsic viscoelasticity of elastomeric skins introduces significant hysteresis, where the deformation signal lags behind the applied force, leading to zero-point drift and measurement inaccuracies. Prior efforts to mitigate these effects primarily focus on classical rheological frameworks, such as the Maxwell or Kelvin-Voigt models \cite{tan1}. For instance, Kulkarni et al. proposed a visco-hyperelastic model to capture the time-dependent stress response and the Mullins effect during cyclic loading \cite{tan2}. Although physically accurate, these differential-equation-based approaches entail significant computational overhead, which is often incompatible with the microsecond-level processing requirements of neuromorphic systems. 

There is a clear need for a computationally efficient, incremental update scheme that aligns with the asynchronous and sparse nature of event data. To this end, we propose a hysteresis-aware update law based on a spatial gain decay mechanism to ensure real-time stability without the need for complex rheological parameter identification.

\section{SpikingTac Sensor Design}

\begin{figure}[!t]
 \centering
 \subfloat[]{
  \includegraphics[width=0.8\columnwidth]{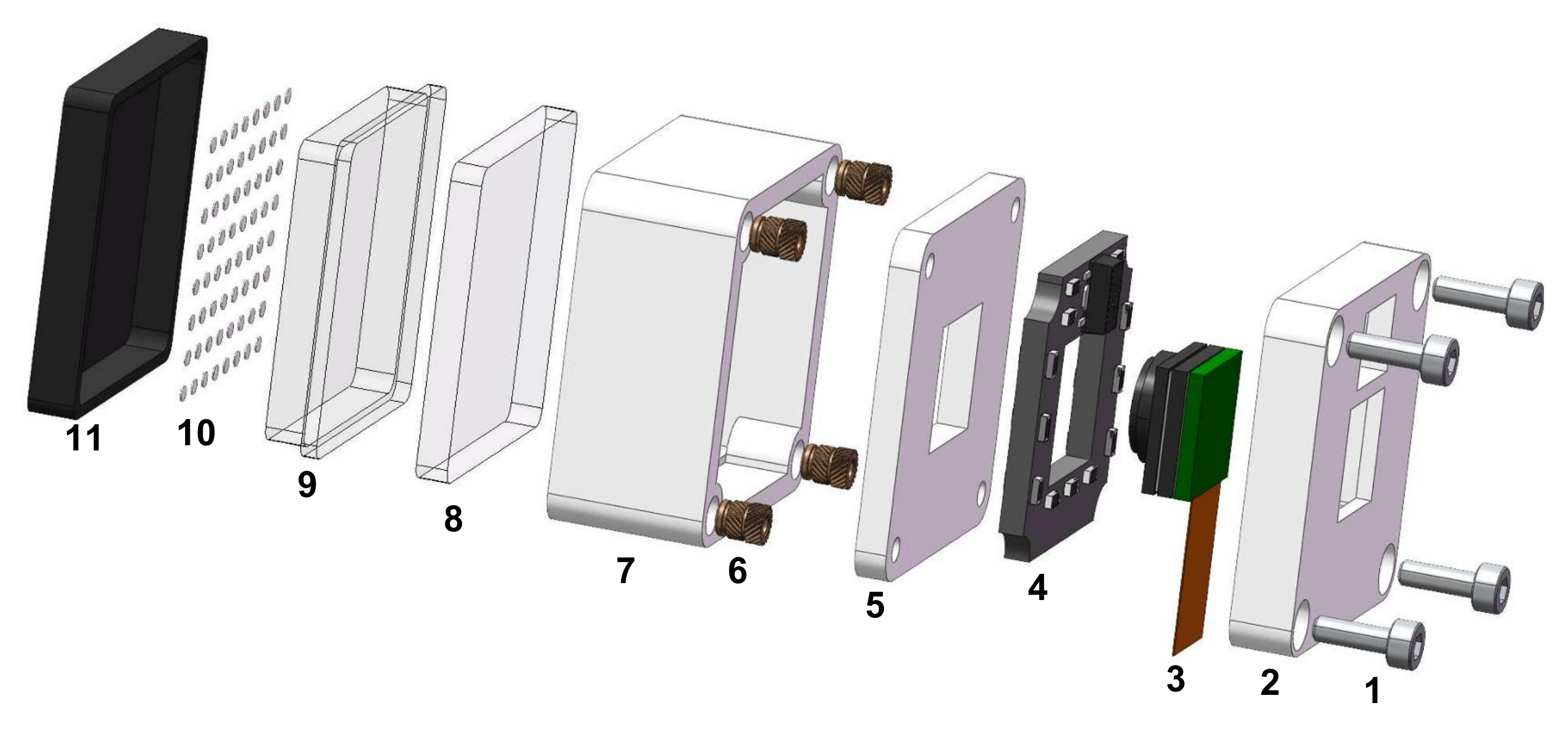}
  \label{fig1:a}
 }\\ 
 \subfloat[]{
  \includegraphics[width=0.8\columnwidth]{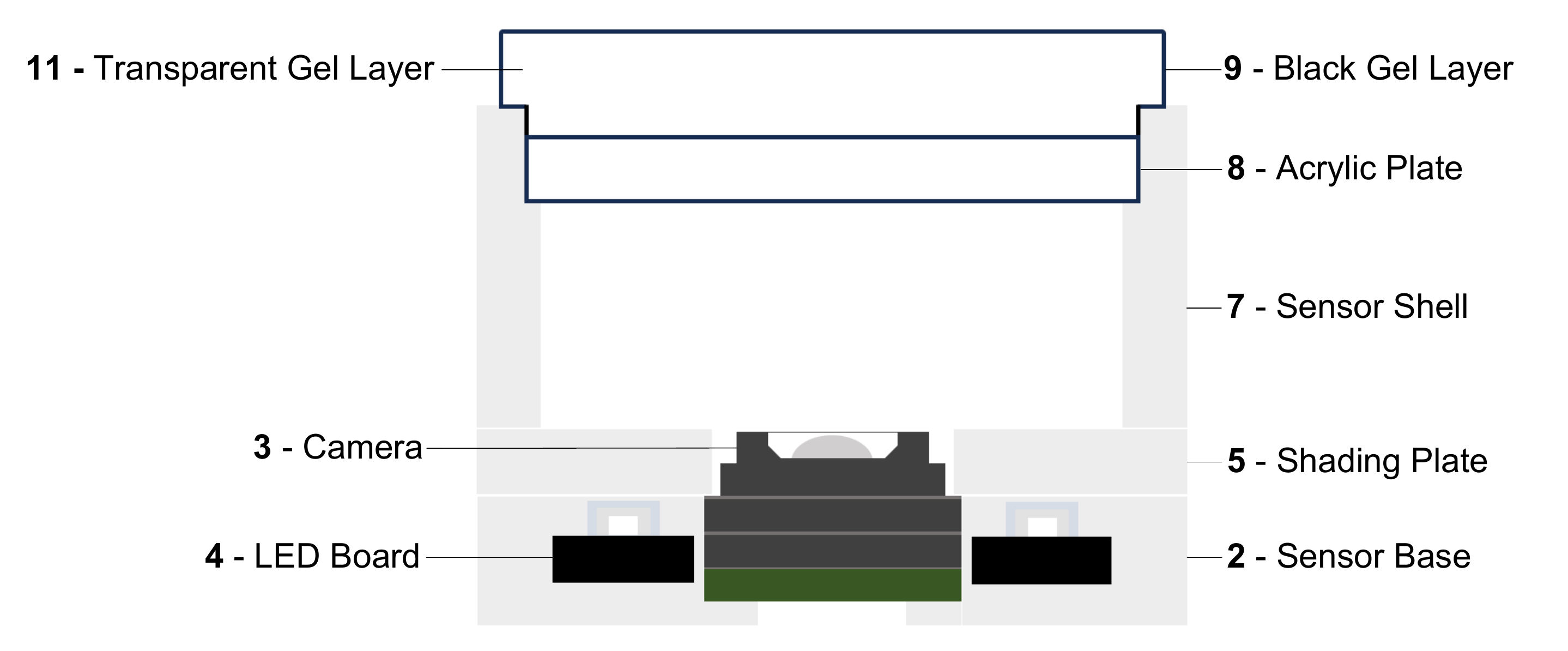}
  \label{fig1:b}
 }
 \caption{Design of SpikingTac. (a) The exploded view of SpikingTac. The components labeled as 1 are screws for connecting the sensor base, the shading plate and the sensor
shell. The components labeled as 6 are heat-set threaded inserts. The components labeled as 10 are 64 white circular markers arranged in an 8x8 grid.
(b) The schematic diagram of SpikingTac.}
 \label{fig1}
\end{figure}

We introduce SpikingTac, a compact, event-based optical tactile sensor designed for accessibility, high temporal resolution, and seamless integration into dexterous robotic hands. The design prioritizes a minimalist structural configuration to minimize specialized expertise and footprint while maximizing SNR.

\begin{figure}[!t]
 \centering
  \includegraphics[width=1.0\columnwidth]{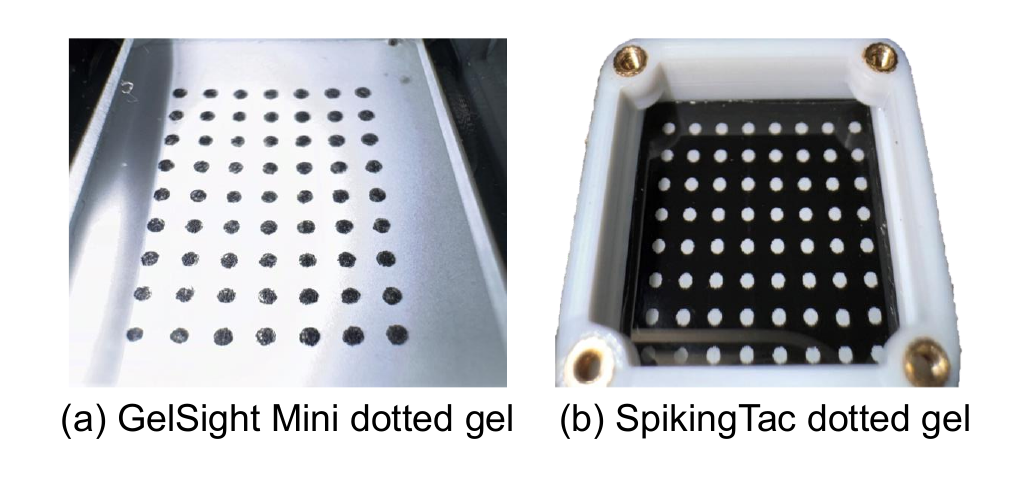}%
 \caption{Comparison image of Gelsight Mini dotted gel and SpikingTac dotted gel.}
 \label{marker}
\end{figure}

\subsection{Hardware}
Fig. \ref{fig1} illustrates the components of SpikingTac. The sensor's foundation is a 3D-printed base and shell, providing high dimensional stability and impact resistance. The internal illumination system utilizes a constant-current LED driver and a shading plate to provide uniform, flicker-free light, ensuring the event-based camera operates with minimal background noise. The sensing interface consists of a transparent silicone gel layer bonded to an acrylic window for high-sensitivity deformation. On top, a tactile perception layer, which is made of opaque black silicone, shields the camera from ambient light. This layer is embedded with a 8$\times$8 array of white markers using a durable cold-transfer printing technique, which provides clearer marker patterns that are more resistant to tearing, as shown in Fig. \ref{marker}. By capturing the displacement of these markers as discrete event streams, SpikingTac achieves microsecond-level temporal resolution.

\subsection{Design Advantages}

As highlighted in Table \ref{tab1}, SpikingTac differentiates itself through its extreme compactness and robustness. With a volume of only $9196\text{ mm}^3$, it is approximately 13\% the size of the Evetac sensor, offering the lowest dimension-to-sensing area (D/A) ratio among its peers. This miniature form factor allows for versatile mounting on various robotic end-effectors. Furthermore, the opaque sensing interface ensures the sensor is agnostic to external lighting conditions, focusing purely on high-frequency, real-time tactile transients with minimal latency.

\section{Methodolog}

\begin{figure}[!t]
 \centering
  \includegraphics[width=0.9\columnwidth]{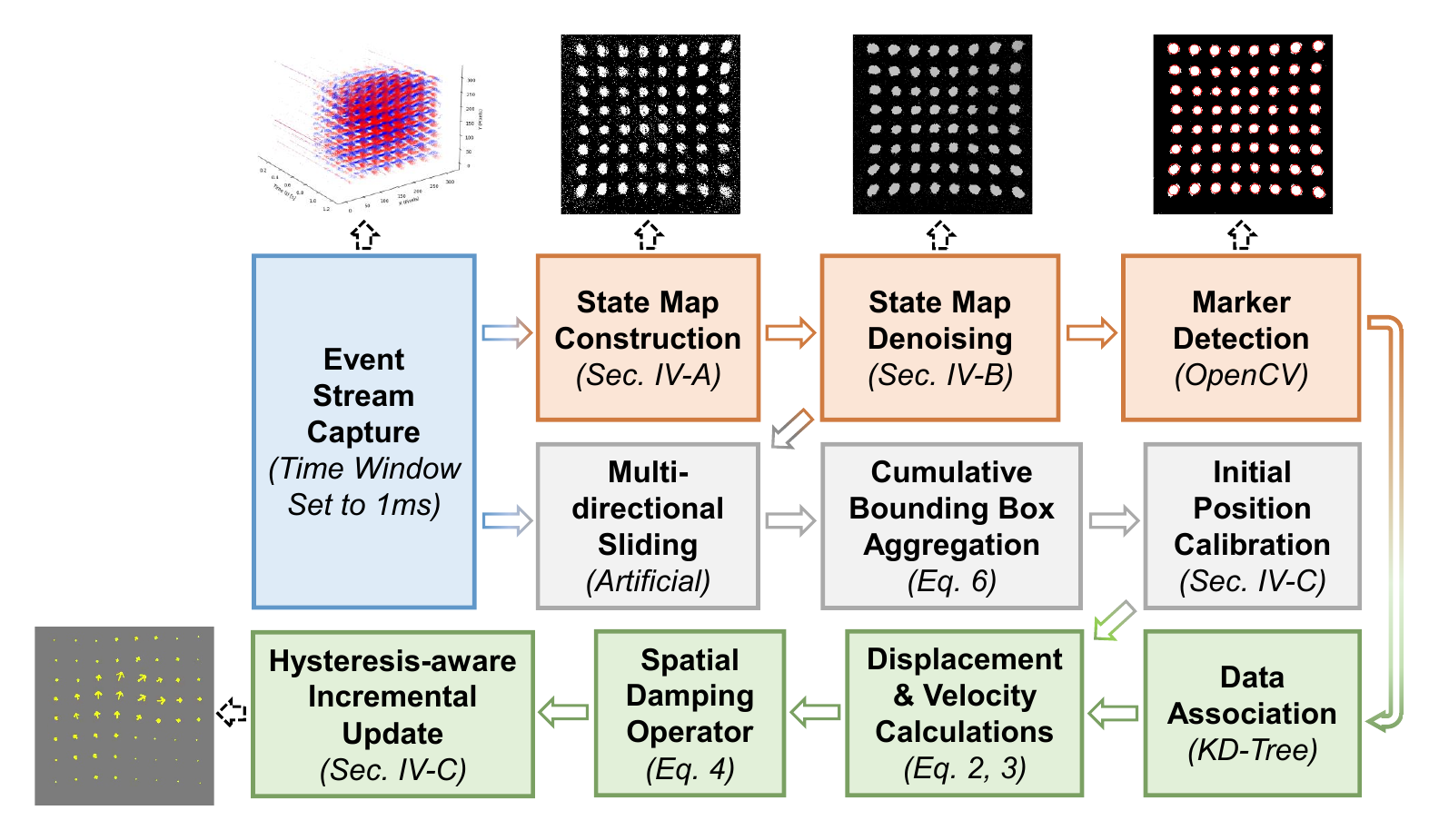}
 \caption{Overview of the proposed event-driven marker tracking pipeline. The process begins with Event Stream Capture, followed by three core phases: Geometric Reconstruction (orange), which builds and denoises the state map; Robust Calibration (grey), which aggregates multi-directional motion to determine initial reference positions; and Dynamic Tracking (green), which implements a hysteresis-aware update law with a spatial damping operator to ensure zero-baseline stability.}
 \label{method_pipeline}
\end{figure}

The proposed tracking pipeline transforms asynchronous event streams into high-fidelity displacement fields through a three-stage architecture, as illustrated in Fig. \ref{method_pipeline}.

\subsection{Event-Driven State Map Construction}

An event camera triggers asynchronous events $e_i = (x_i, y_i, p_i, t_i)$ when the log-intensity change at pixel $(x_i, y_i)$ exceeds a threshold, where $p_i \in \{+1, -1\}$ and $t_i$ is the microsecond-level timestamp.

\begin{figure}[!t]
 \centering
  \includegraphics[width=0.9\columnwidth]{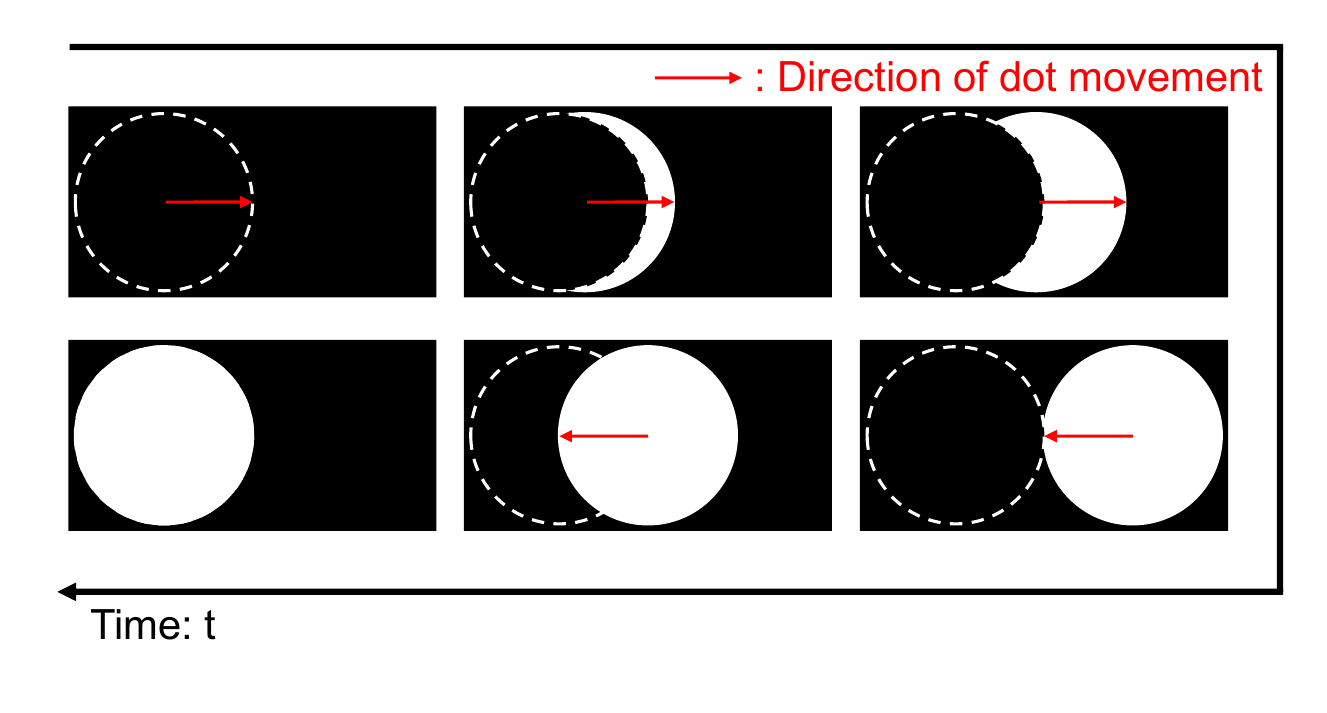}%
 \caption{Conceptual illustration of the event-driven marker reconstruction. Complete shape reconstruction: As the white circular marker translates by a distance equal to its diameter, the asynchronous event stream (represented by $p=+1$ and $p=-1$ polarities) cumulatively populates and clears the state map $M(x,y)$, eventually reconstructing the full geometry at the terminal position.}
 \label{state map}
\end{figure}

To recover marker geometry from temporal contrast, we define a Dynamic State Map $M(x, y)$ as a persistent spatial memory. Conventional frame-based integration is discarded in favor of an asynchronous update law. As a marker translates, its leading edge generates positive events ($p=+1$) by increasing local intensity, while its trailing edge generates negative events ($p=-1$). As illustrated in Fig. \ref{state map}, the cumulative effect of these polarities over a characteristic displacement (equivalent to the marker diameter) reconstructs the full marker geometry. Once established, this profile remains invariant under continuous motion, as new events concurrently populate leading-edge occupancy and clear trailing-edge vacancies.

Formally, $M$ is initialized as a null matrix and updated for each incoming event $e_i$ according to:

\begin{equation}
M(x_i, y_i) = 
\begin{cases} 
1 \text{ (White)}, & \text{if } p_i = +1 \\
0 \text{ (Black)}, & \text{if } p_i = -1 
\end{cases}
\label{eq_state_map}
\end{equation}

This mechanism ensures that $M(x, y)$ functions as a binary occupancy grid, where positive events ``paint'' the marker's advancing footprint and negative events ``erase'' its vacated path, maintaining an up-to-date representation of the tactile interface's internal state.

\subsection{Event-Driven State Reconstruction and Denoising}

As the global state map $M$ is initially null, a Bootstrap Calibration phase is conducted where multi-directional sliding motions populate the markers' reference positions. This self-calibrating approach accounts for fabrication tolerances by defining the sensor's neutral state through empirical observation rather than rigid geometric priors.

To suppress background activity--modeled as a spatio-temporally isolated Poisson-point process \cite{v2e}--we employ an unsupervised denoising framework. Following the architecture in \cite{evflownet} and the loss formulations in \cite{neutac}, a U-Net based encoder-decoder is utilized to map the raw, noisy state map into a high-contrast representation. The network is trained on a diverse dataset of pressing, rotating, and sliding interactions, ensuring generalization across complex contact states. The denoised output is then binarized, and marker centroids are extracted using a Blob Detection algorithm.

To maintain temporal consistency under extreme strain, where markers may exhibit morphological incompleteness or structural gaps, we implement a k-d tree-based data association strategy. For each asynchronous update, the k-d tree is queried to match newly detected centroids with identities from the preceding state within a proximity threshold. If a marker candidate fails to meet the detection criteria due to low local contrast, its position is temporally sustained via a zero-order hold from its last known state. This mechanism ensures a continuous and persistent 64-point displacement field throughout high-dynamic manipulation.

\subsection{Hysteresis-aware Incremental Update}

\begin{figure}[!t]
    \centering
    \includegraphics[width=0.55\linewidth]{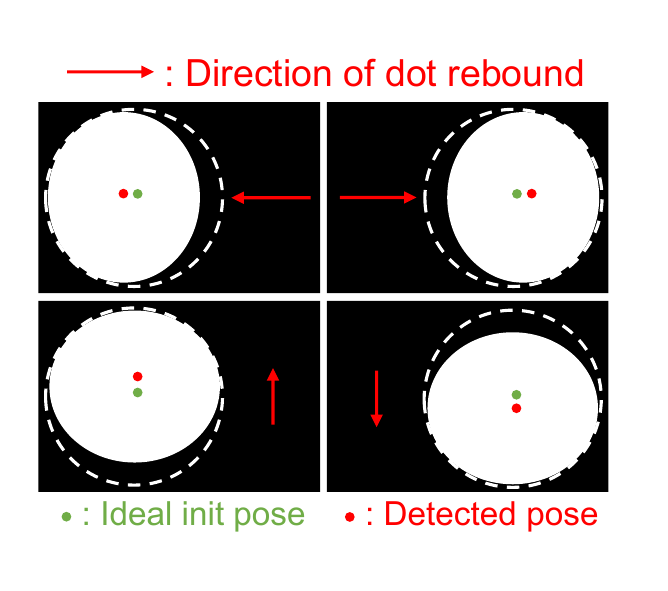}
    \caption{Illustration of the systematic offsets during the marker restoration phase. Due to the viscoelastic hysteresis of the silicone elastomer, the detected centroid $P_{det}$ (red dot) fails to perfectly realign with the ideal initial reference $P_{init}$ (green dot) once the external force is removed, resulting in a consistent residual error. The red arrows indicate the rebound direction of the markers.}
    \label{fig4}
\end{figure}

Despite robust data association, detected centroids $P_{det,t}$ exhibit systematic offsets during the restoration phase. This residual error, often observed upon unloading, stems from the coupling of elastomeric viscoelastic hysteresis \cite{tan3} and the intrinsic spatio-temporal latency of event-driven flow, as shown in Fig. \ref{fig4}. To ensure zero-point stability, we propose a Spatial Damping Operator based on an incremental state evolution model.

We maintain a persistent state $P_{real,t}$ that evolves according to the instantaneous velocity vector while attenuating near-field jitter. Let $P_{init}$ denote the calibrated reference coordinates. We define the instantaneous displacement $\vec{a}_t$ and the velocity increment $\vec{b}_t$ as:

\begin{equation}
    \vec{a}_t = P_{det, t} - P_{init}
\end{equation}
\begin{equation}
    \vec{b}_t = P_{det, t} - P_{det, t-1}
\end{equation}

The refined position is updated via a distance-dependent gain factor $\Gamma(d)$:

\begin{equation}
 P_{real, t} = P_{real, t-1} + \Gamma(\lVert \vec{a}_t \rVert) \cdot \vec{b}_t
\end{equation}where the gain function $\Gamma(d)$ is defined as:
\begin{equation}
    \Gamma(d) =
    \begin{cases}
        1.0, & \text{if } d \ge \delta \\
        \gamma, & \text{if } d < \delta
    \end{cases}
    \label{eq_damping}
\end{equation}

The threshold $\delta$ defines an uncertainty window around the equilibrium position. By setting a damping coefficient $\gamma < 1$ within this window (e.g., $\delta=4.0, \gamma=0.7$), the algorithm effectively absorbs overshoots caused by material lagging and filters high-frequency jitter. This ensures the markers converge smoothly to $P_{init}$, providing a stable baseline for force estimation.

\subsection{Robust Initial Position Calibration}
\label{4d}

To establish a precise reference $P_{init}$ resistant to residual strains and elastic memory, we implement a Multi-directional Bounding Box Aggregation method. During calibration, the sensor surface is subjected to diverse sliding interactions. For each marker $i$, the system maintains a cumulative bounding box $\mathcal{B}_i = [x_{min}, y_{min}, x_{max}, y_{max}]_i$, which is iteratively updated to encompass the spatial extent of the marker across all equilibrium states.

The final calibrated position is defined as the geometric center of the aggregated bounding box:

\begin{equation}
    P_{init, i} = \left( \frac{x_{min} + x_{max}}{2}, \frac{y_{min} + y_{max}}{2} \right)
\end{equation}

This approach defines $P_{init}$ as the mechanical ``neutral point'', minimizing initial bias induced by varying contact histories or fabrication pre-strains.

\section{Experiments}
In this section, we present three experiments to showcase SpikingTac's properties. First, we evaluate the robustness of our point-tracking algorithm under complex, high-dynamic interactions. Second, we investigate the sensor's high-bandwidth response through a high-speed collision detection task. Finally, we demonstrate SpikingTac's precision in a multi-scale circular geometry estimation task using active tactile exploration. These experiments collectively validate the sensor's superior temporal resolution, tracking stability, and accuracy in demanding robotic applications.

\subsection{Point Tracking}

This experiment investigates the robustness of the proposed tracking algorithm under complex, high-dynamic tactile interactions. As noted in \cite{evetac}, a critical challenge for event-based tactile sensors is the high volume of interference events triggered at object boundaries during rapid contact. These ``edge-induced'' events often contradict the fundamental assumption that event streams originate solely from marker motion, thereby leading to significant tracking drift. In contrast, the SpikingTac sensor benefits from an optimized optical interface and surface characteristics that suppress such interference events during sliding. Consequently, the event stream remains highly consistent with the actual displacement of the markers, providing a high-fidelity signal source that fundamentally enhances tracking reliability.

\subsubsection{Setups}

\begin{figure}[!t]
    \centering
    \includegraphics[width=1.0\columnwidth]{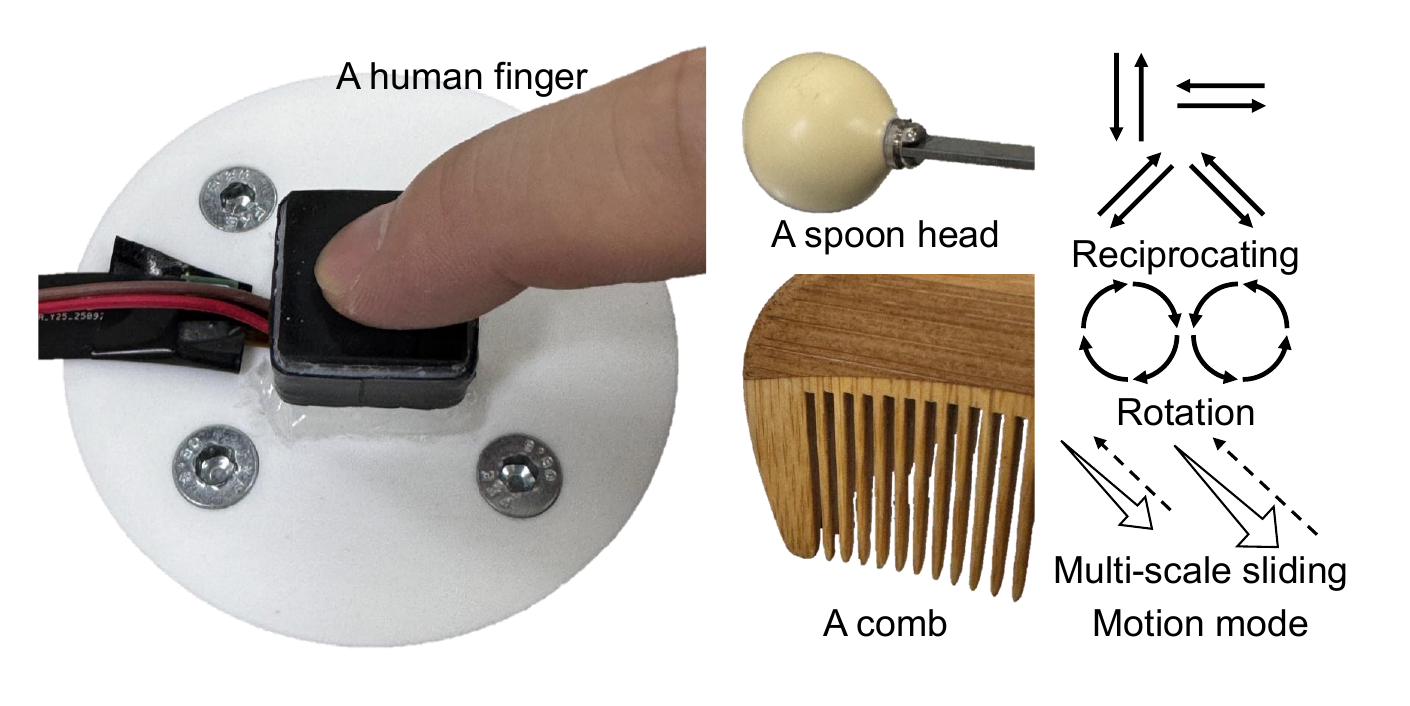}
    \caption{Experimental setup for point tracking and various contact interaction modes during data acquisition.}
    \label{exp1_setup}
\end{figure}

\begin{figure}[!t]
    \centering
    \includegraphics[width=0.9\columnwidth]{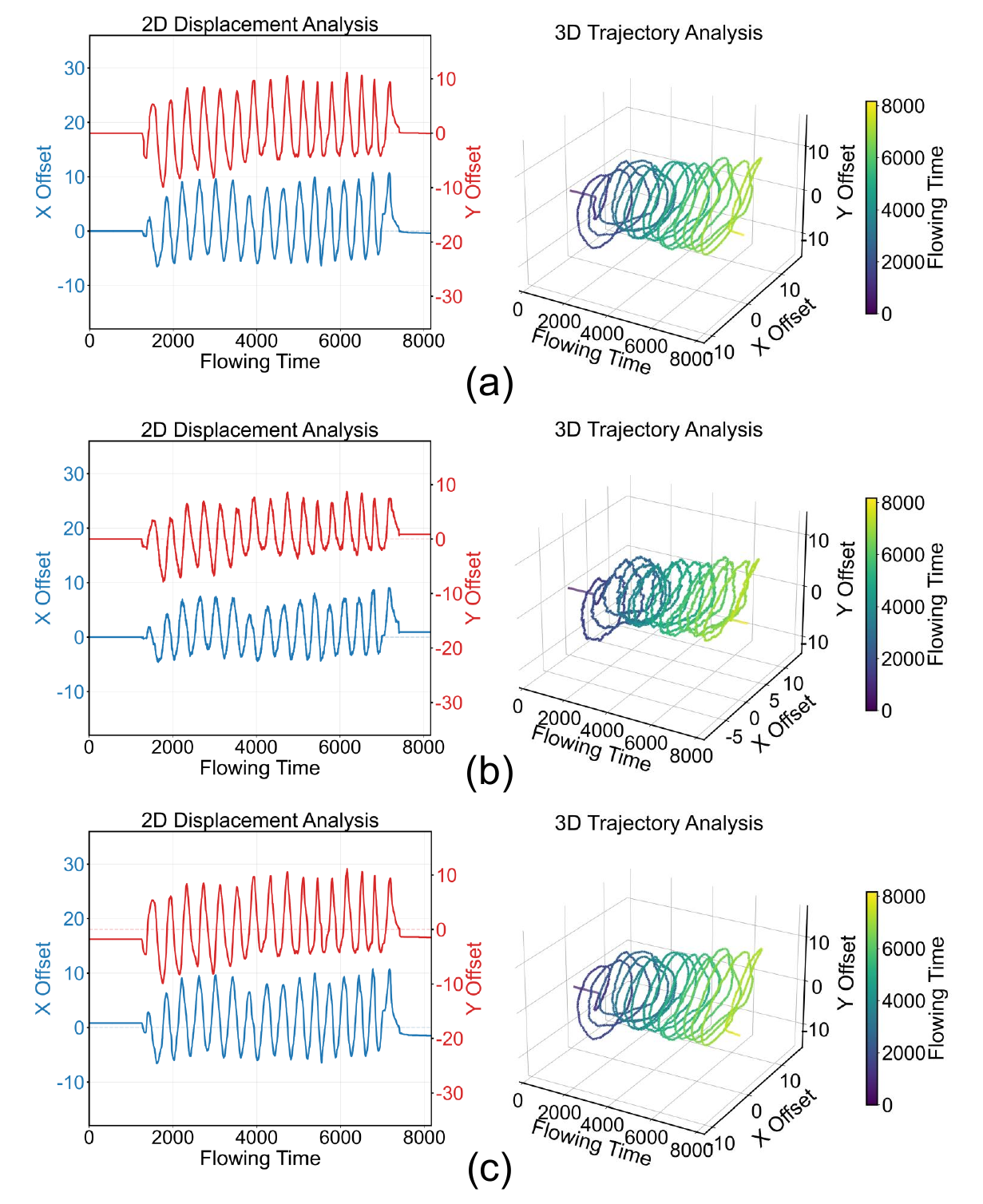}
    \caption{Tracking stability characterization of a single keypoint under repetitive contact tasks. (a) Ours; (b) Evetac; (c) NeuTac.}
    \label{exp1_process}
\end{figure}

The experimental platform involves the SpikingTac sensor fixed on an optical bench. We performed ten distinct contact trajectories using diverse objects, including a human finger, a spoon head, and a comb, as shown in Fig. \ref{exp1_setup}. Our approach is benchmarked against two state-of-the-art baselines: the regularized point tracking method from \cite{evetac} and the denoising-based detection method from \cite{neutac}.

To quantitatively evaluate tracking performance, we analyze end-point consistency. Since all trajectories begin and end in a non-contact state, markers should theoretically return to their initial equilibrium positions. Performance is assessed using two metrics:
\begin{enumerate}
    \item Success Rate: The percentage of trajectories where markers return to within a 5-pixel radius of their initial positions. Given the $320 \times 320$ resolution, this criterion is significantly more stringent than the 20-pixel threshold typically employed in previous studies.
    \item Deviation Metrics: The mean distance error and standard deviation across all markers in the final state, calculated relative to the ground-truth initial positions.
\end{enumerate}


\subsubsection{Implementation Details}
Departing from conventional studies focused on simple shear or translational motions, our protocol incorporates torsion, rapid repetitive sliding, and circular movements to rigorously evaluate stability. Event data were recorded at a temporal resolution of 1000~Hz. To ensure unbiased evaluation, tracking was not performed in real-time during the physical interactions; instead, the raw event data were collected and subsequently processed through each algorithm to evaluate their respective tracking performance offline. Without loss of generality, a representative trajectory of a single contact point during the interaction process is extracted, as illustrated in Fig. \ref{exp1_process}.

\subsubsection{Results}
The tracking success rates are summarized in Table \ref{tab2}. Both our method and NeuTac achieved a 100\% success rate. In contrast, Evetac failed in two out of ten complex interactions. Analysis of these failures indicates that during large-angle torsion, markers near the sensor boundaries were lost, resulting in errors exceeding the 5-pixel threshold. Additionally, extreme drag-and-release actions induced rapid transient motions that caused Evetac to lose its tracking lock.

\begin{figure}[!t]
    \centering
    \includegraphics[width=0.95\columnwidth]{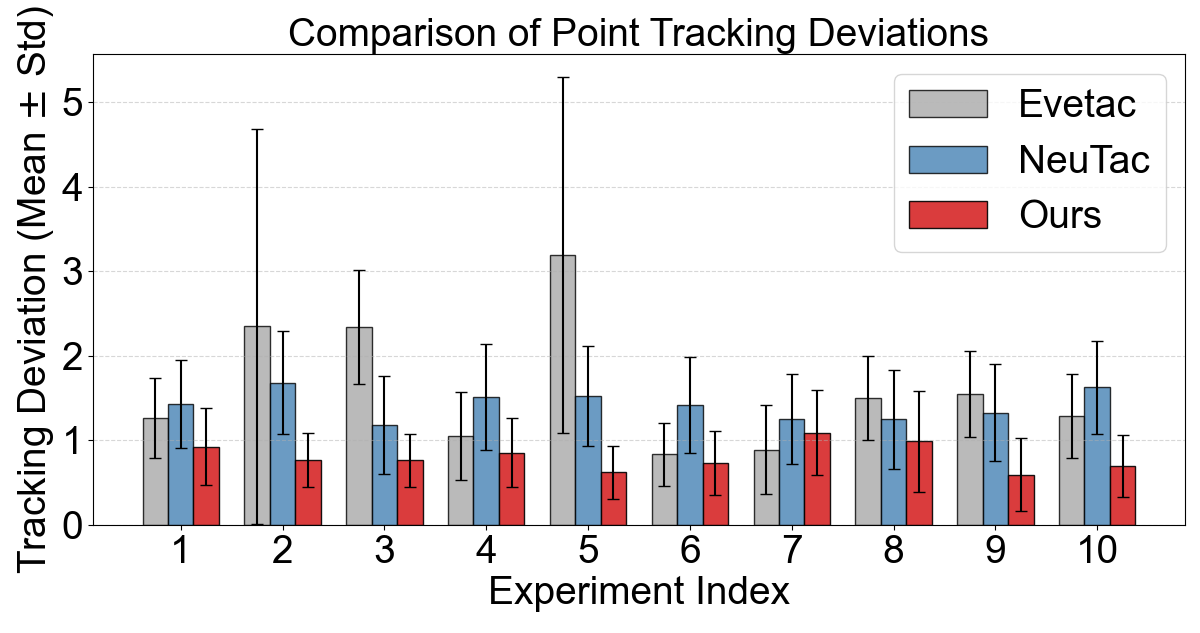}
    \caption{Quantitative comparison of point tracking deviations across ten experimental sequences.}
    \label{fig6}
\end{figure}

Fig. \ref{fig6} illustrates the tracking deviations for all ten trajectories. Although NeuTac maintains a high success rate, it exhibits a systematic homodirectional regression bias, where markers consistently drift in the direction of motion, leading to higher offset errors. Evetac demonstrates lower stability with fluctuating deviations in its successful trials. Notably, Evetac achieved the minimum drift error in Trajectory 7, which involved localized circular finger motions. Such patterns generate specific ``flow'' events within confined regions that Evetac's regression model is well-suited to handle. However, from a holistic perspective, our method consistently achieved the lowest mean deviation (0.8039) and standard deviation (0.4097), demonstrating superior robustness and stability across the broadest range of complex interaction scenarios.

\begin{table}[!t]
	\renewcommand{\arraystretch}{1.3}
	\caption{Performance Comparison of Point Tracking Algorithms.}
	\centering
	\label{tab2}
	\resizebox{\columnwidth}{!}{
		\begin{tabular}{l c c c}
			\hline\hline \\[-3mm]
			\multicolumn{1}{c}{Algorithm} & \multicolumn{1}{c}{Success Rate} & \multicolumn{1}{c}{Average Mean} & \multicolumn{1}{c}{Average Std} \\[1.6ex] \hline
			Evetac \cite{evetac} & 80\% & 1.6276 & 0.8525 \\
			NeuTac \cite{neutac} & \textbf{100\%} & 1.4212 & 0.5742 \\ 
			\textbf{Ours} & \textbf{100\%} & \textbf{0.8039} & \textbf{0.4097} \\ [1.4ex]
			\hline\hline
		\end{tabular}
	}
\end{table}

\subsection{Dynamic Collision Detection via High-Frequency Spiking Sensing}
\label{5b}

We explore the high-frequency sensing capabilities of SpikingTac through a dynamic collision detection task. Event-based tactile sensors are inherently sensitive to dynamic stimuli, as transient physical interactions trigger a rapid surge in event streams \cite{fast sensing}. To evaluate this high-frequency capability, we utilize the raw event count as the primary feature for detection. By defining a cumulative temporal window $\Delta t = 1\,\text{ms}$, the sensor facilitates a perception frequency of $1000\,\text{Hz}$ for transient signal detection.

\subsubsection{Setups}
\begin{figure}[!t]
    \centering
    \subfloat[]{
        \includegraphics[width=0.45\linewidth, angle=0]{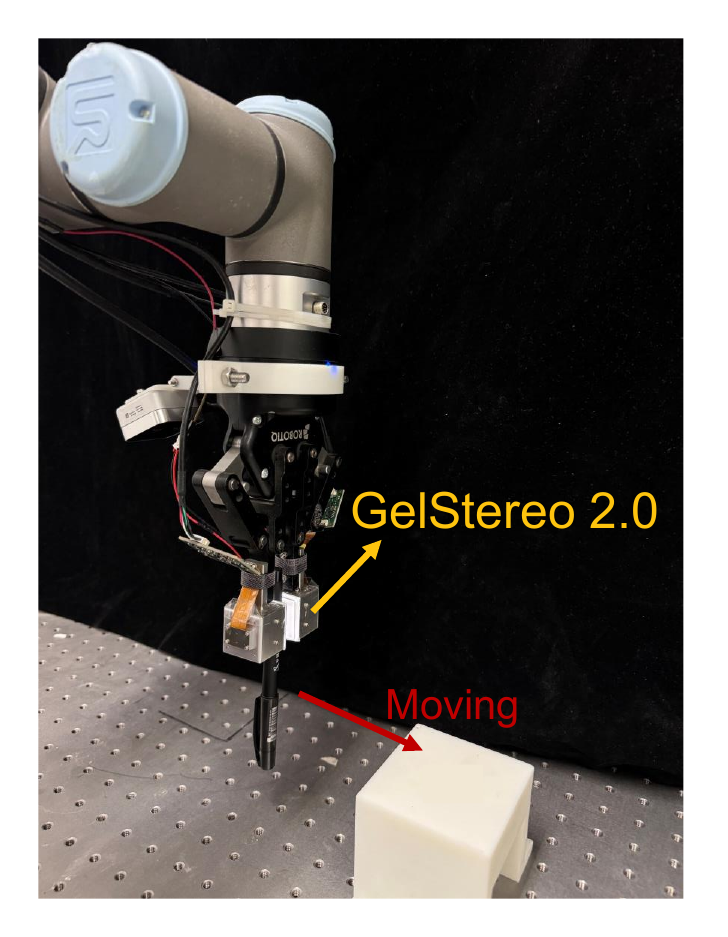}
        \label{fig7a}
    }
    \subfloat[]{
        \includegraphics[width=0.45\linewidth, angle=0]{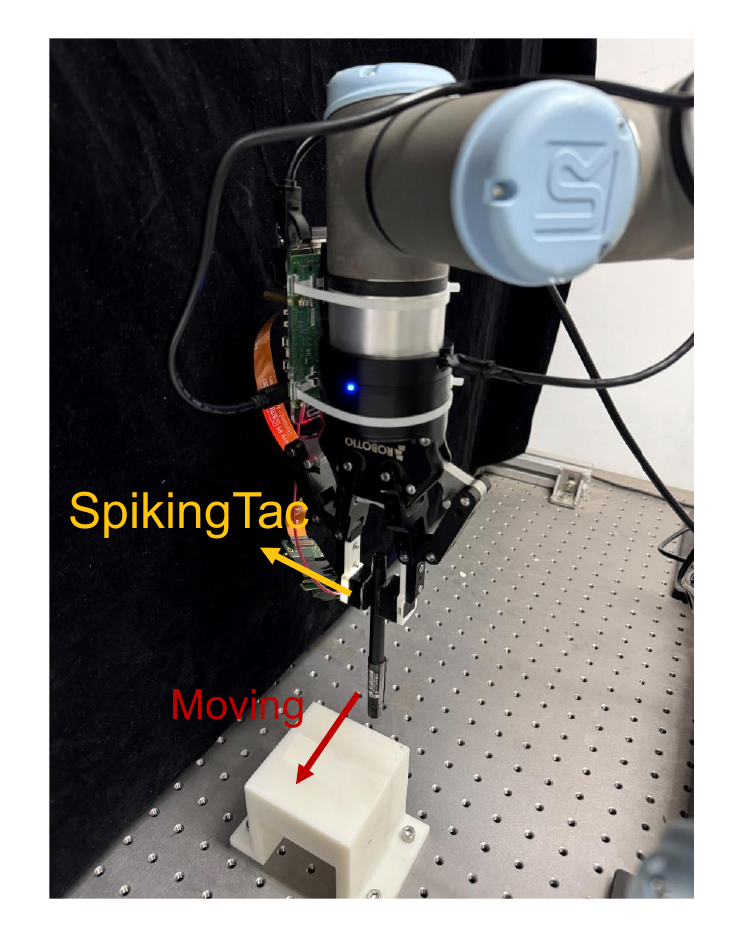}
        \label{fig7b}
    }
    \caption{Experimental setup for collision detection. (a) GelStereo 2.0; (b) Ours.}
    \label{fig7}
\end{figure}

The goal of this task is to evaluate the latency and accuracy of the proposed method during high-speed contact. As illustrated in Fig. \ref{fig7}, both the SpikingTac sensor and a frame-based baseline (GelStereo 2.0 \cite{gelstereo2.0}) are integrated into a Robotiq-85 gripper mounted on a UR3 industrial robot. A probe is vertically grasped and driven along a linear trajectory at constant velocities $v \in [0.01, 0.18]\,\text{m/s}$ toward a rigid obstacle.

To assess performance, three critical Cartesian poses are recorded:
\begin{itemize}
    \item $\mathbf{X}_{real}$: The ground-truth contact pose obtained via manual teleoperation until the onset of physical contact.
    \item $\mathbf{X}_{collision}$: The robot end-effector pose at the instant the collision signal is triggered.
    \item $\mathbf{X}_{stop}$: The final equilibrium pose after the robot reaches a complete standstill.
\end{itemize}
We define the detection deviation as $\Delta \mathbf{X}_{p} = \mathbf{X}_{collision} - \mathbf{X}_{real}$ and the post-impact overshoot as $\Delta \mathbf{X}_{e} = \mathbf{X}_{stop} - \mathbf{X}_{real}$.

\begin{figure}[!t]
 \centering
  \includegraphics[width=0.9\columnwidth]{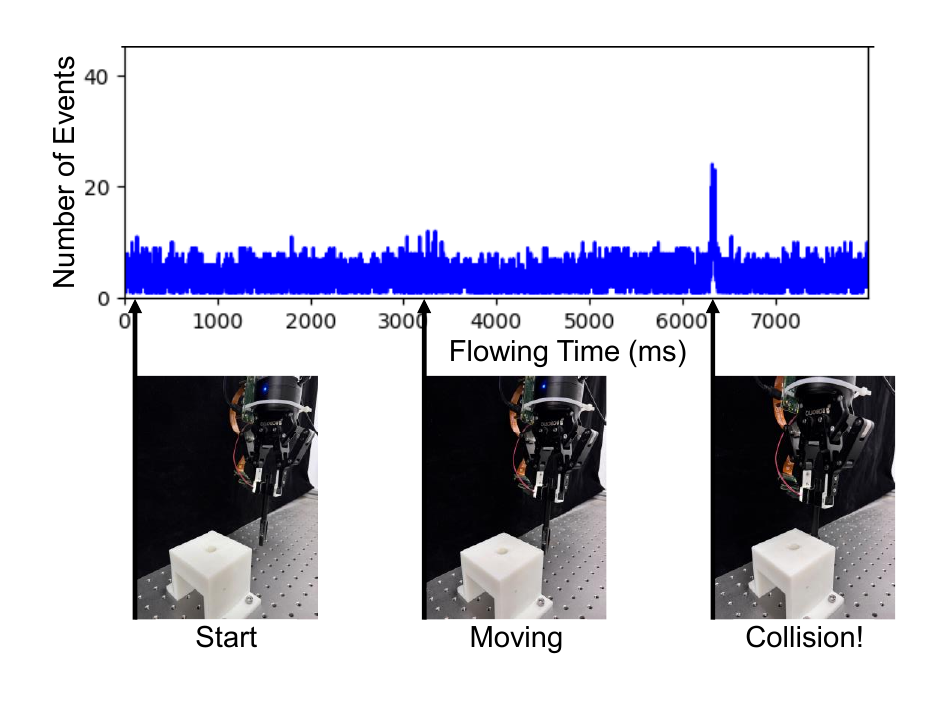}%
 \caption{The experimental process of collision detection.}
 \label{collision_process}
\end{figure}

\subsubsection{Implementation Details}
For the SpikingTac sensor, we define a cumulative temporal window $\Delta t = 1\,\text{ms}$, facilitating a perception frequency of $1000\,\text{Hz}$. A collision is registered when the event count within $\Delta t$ exceeds a predefined threshold.

For the frame-based GelStereo 2.0 baseline, a displacement-based heuristic is implemented. Let $\mathbf{P}_{init} = \{\mathbf{p}_{1}, \dots, \mathbf{p}_{n}\}$ denote the initial coordinates of the markers. The real-time displacement sum $D(t)$ is calculated as:
\begin{equation}
D(t) = \sum_{i=1}^{n} \|\mathbf{p}_{i,cur}(t) - \mathbf{p}_{i,init}\|_2
\end{equation}
A collision is triggered when $D(t) > \epsilon_{threshold}$, where the threshold is determined by the upper bound of the quiescent interval. Upon detection of a collision signal from either sensor, a stop command is immediately issued to the UR3 controller.

\begin{figure}[!t]
 \centering
  \includegraphics[width=0.98\columnwidth]{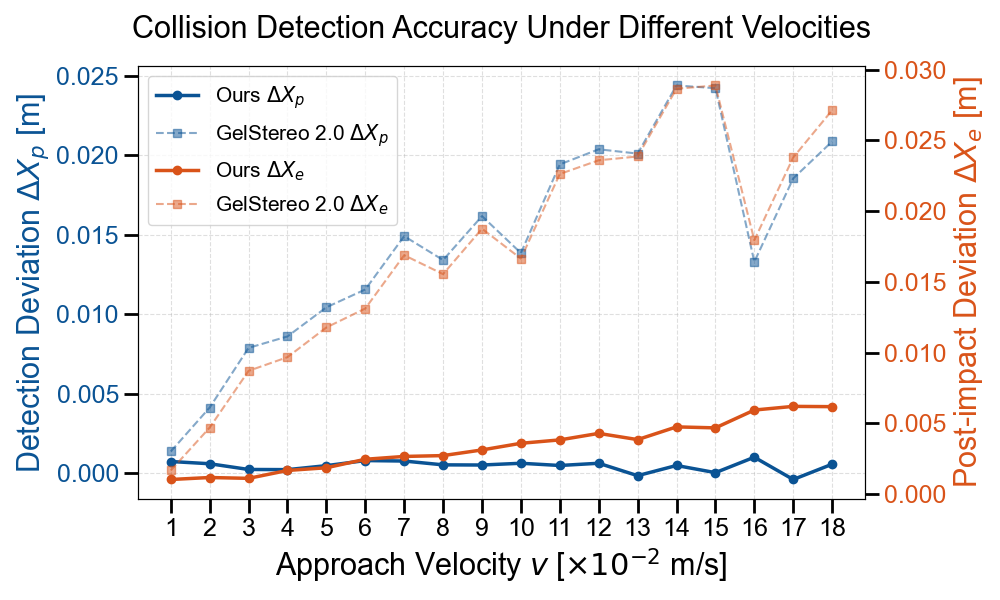}%
 \caption{Quantitative evaluation of collision detection performance under varying approach velocities. The left y-axis indicates the detection deviation ($\Delta X_p$) at the instant the collision signal is triggered, while the right y-axis represents the post-impact overshoot ($\Delta X_e$) after the robot reaches a complete standstill. Solid lines denote our SpikingTac sensor, and dashed lines represent the frame-based GelStereo 2.0 baseline \cite{gelstereo2.0}.}
 \label{fig8}
\end{figure}

\subsubsection{Results}
Fig. \ref{collision_process} illustrates the temporal evolution of the event count during the UR3 manipulator's trajectory. Upon the occurrence of a collision, a pronounced spike in the event rate is observed, triggering the collision detection algorithm and the subsequent issuance of an immediate stop command to the robot.

Fig. \ref{fig8} summarizes the comparative results between SpikingTac and GelStereo 2.0 across varying approach velocities. As shown in the left y-axis of Fig. \ref{fig8}, SpikingTac maintains a near-constant sub-millimeter deviation $\Delta X_p$ (mean $\approx 0.53\,\text{mm}$), demonstrating a velocity-independent response. In contrast, the frame-based baseline exhibits a significant linear growth in $\Delta X_p$ as velocity increases, reaching a maximum deviation of approximately $24.3\,\text{mm}$. This suggests that the proposed sensor possesses a much higher sampling bandwidth and lower latency, ensuring that the collision signal is triggered almost instantaneously upon physical contact.

Several data points for $\Delta X_p$ exhibit minor negative values, primarily attributed to stochastic resetting errors and geometric inconsistencies. Specifically, micro-slips at the contact interface during the experimental reset phase prevent a perfectly consistent initial alignment of the probe. It is worth noting that these negative offsets are exceptionally small (within $0.5\,\text{mm}$), and their presence underscores the high sensitivity and transparency of SpikingTac. Unlike conventional methods that employ heavy signal filtering which masks such micro-scale variations, our sensor directly captures the true physical interaction, including these inevitable experimental perturbations.

The right y-axis depicts the stopping displacement $\Delta X_e$, which reflects the system's reaction time and braking efficiency. SpikingTac restricts the maximum overshoot to within $6.2\,\text{mm}$, providing a five-fold improvement over the baseline method at high velocities. While the baseline's response becomes erratic at $v > 0.15\,\text{m/s}$, our system maintains a stable and predictable response. This underscores the critical importance of $1000\,\text{Hz}$ sensing frequency in ensuring safety during high-speed robotic tasks, where minimizing post-impact displacement is essential.

\subsection{Multi-scale Circular Geometry Estimation}

We conduct a multi-scale circular geometry estimation task to evaluate the capability of SpikingTac in extracting precise geometric features through active exploration. The objective is to estimate the center position and radius of various circular holes from a random starting point within the hole using tactile feedback.

\begin{figure}[!t]
 \centering
  \includegraphics[width=0.99\columnwidth]{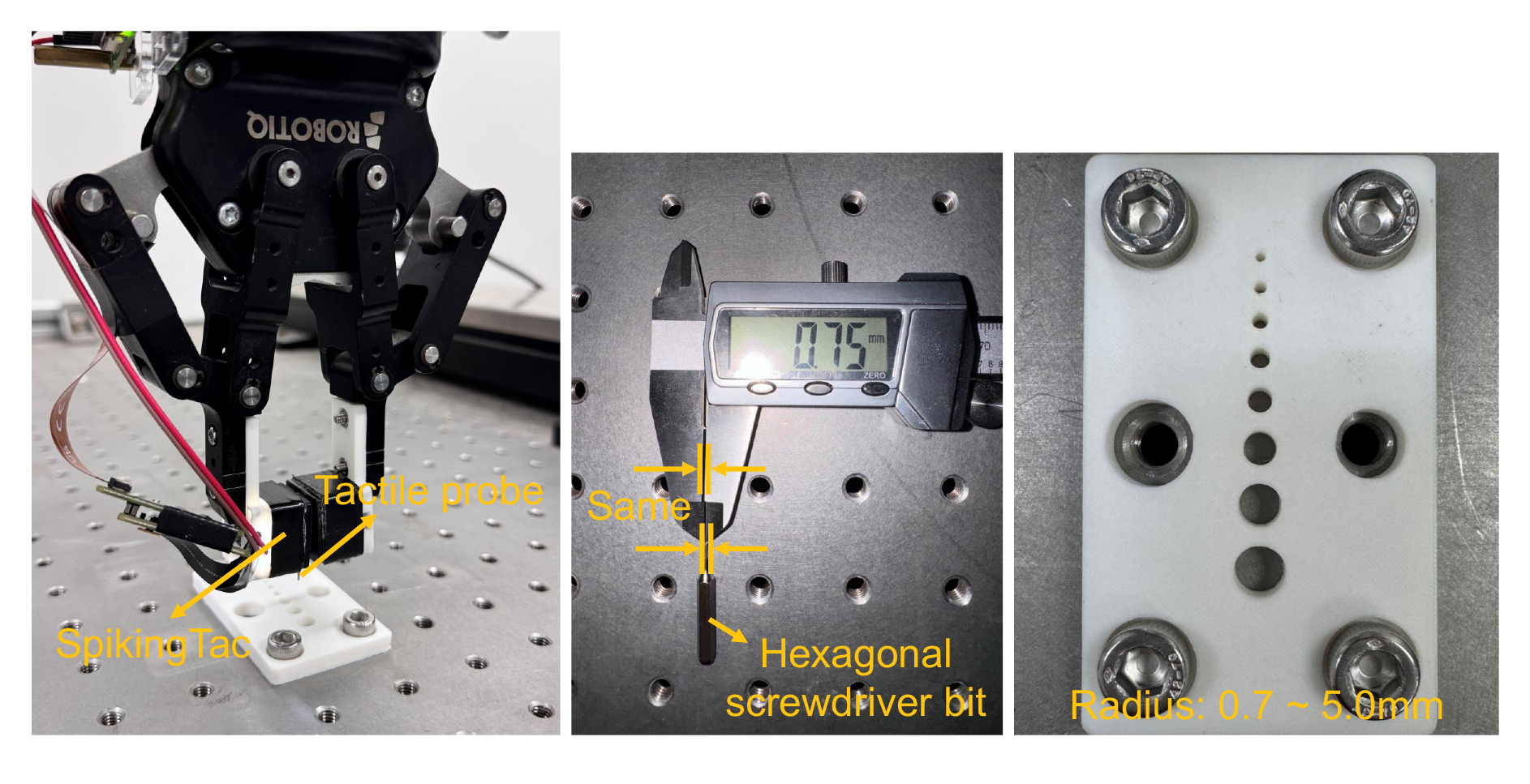}%
 \caption{Experimental setups for multi-scale circular geometry estimation.}
 \label{fig9}
\end{figure}

\subsubsection{Setups}
The experimental setups are illustrated in Fig. \ref{fig9}. Two SpikingTac sensors are mounted on the fingertips of a Robotiq 85 gripper, driven by a UR3 robot arm. A small hexagonal screwdriver bit with a diameter of $D_p = 0.75$ mm is employed as a tactile probe and maintained in a vertical orientation. We fabricated a set of circular holes with radii ranging from 0.7 mm to 5.0 mm, corresponding to common industrial screw hole dimensions.

To ensure the generality of the task, the plane of each hole is aligned parallel to the $xy$-plane of the robot base coordinate system. For each trial, an initial human-estimated position of the hole center is provided. A random offset, smaller than the hole radius, is then added to this initial position to simulate the uncertainty of coarse localization. This ensures that the probe starts within the hole but at an arbitrary location, requiring tactile exploration to determine the exact geometry.

\subsubsection{Implementation Details}

\begin{figure}[!t]
 \centering
  \includegraphics[width=0.9\columnwidth]{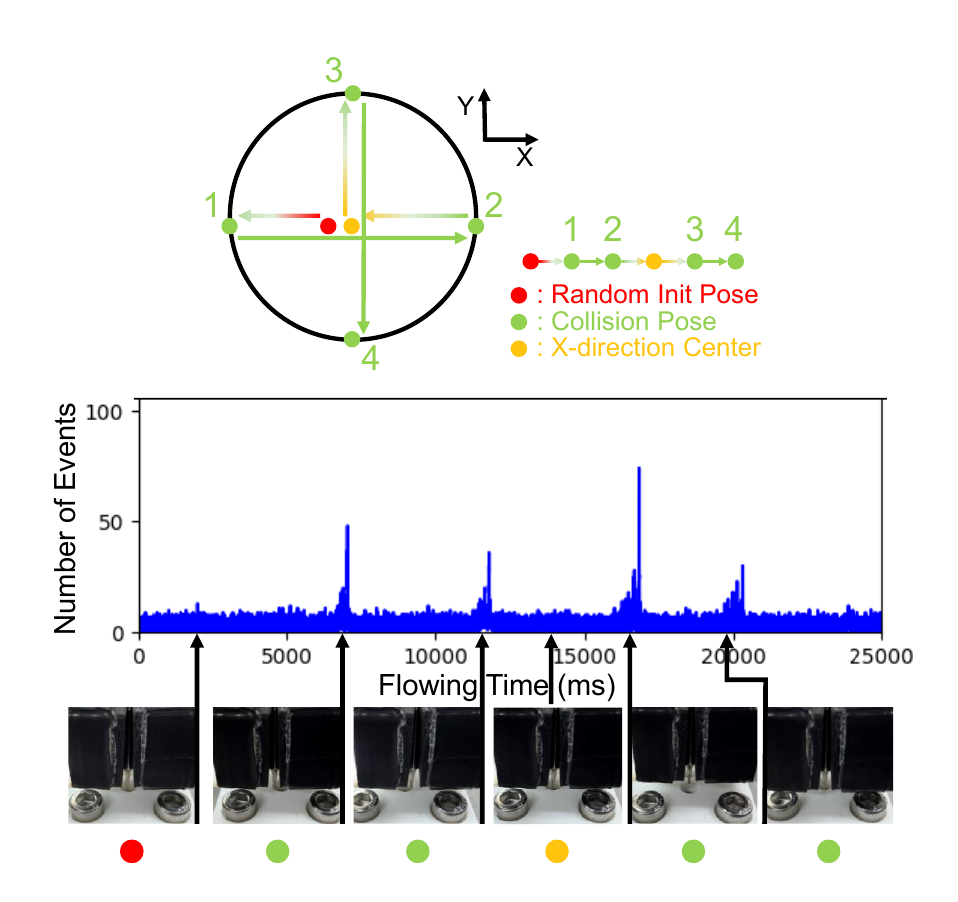}%
 \caption{Schematic of the tactile cross-search strategy for hole center and radius estimation. The probe moves along the $X$ and $Y$ axes sequentially to identify boundary contact points.}
 \label{fig10}
\end{figure}

We implement a cross-search strategy to estimate the geometry. For each hole, the UR3 moves the probe at a constant speed of 0.8 mm/s. Every effective collision contact point is identified using the collision detection method described in Sec. \ref{5b}.

Starting from an initial point $P_0(x_0, y_0)$, the probe first moves along the $+X$ and $-X$ directions to record contact points $P_1(x_1, y_1)$ and $P_2(x_2, y_2)$. The estimated $X$-center is calculated as $x_c = (x_1 + x_2)/2$. The probe is then repositioned to $(x_c, y_0)$ and performs moves along the $+Y$ and $-Y$ directions to capture $P_3(x_3, y_3)$ and $P_4(x_4, y_4)$, yielding the $Y$-center $y_c = (y_3 + y_4)/2$, as shown in Fig. \ref{fig10}.

To bridge the gap between the measured trajectory and the physical geometry, a compensation model is established. Since the sensor records the pose of the probe's central axis, an equivalent radius offset $R_{offset}$ is considered:
\begin{equation}
    R_{offset} = \frac{D_p}{2} = 0.375 \text{ mm}
\end{equation}
Furthermore, we define a critical contact displacement $\delta_{crit}$ as the distance required to trigger the spike threshold from the moment of initial physical contact. In our experiments, $\delta_{crit}$ is set to 0.4 mm. The final estimated hole radius $R_{real}$ is calculated as:
\begin{equation}
    R_{real} = R_{measured} + R_{offset} - \delta_{crit}
\end{equation}
where $R_{measured}$ is the radius derived from the distance between the calculated center and the contact points.

\subsubsection{Results}

\begin{figure}[!t]
 \centering
  \includegraphics[width=0.8\columnwidth]{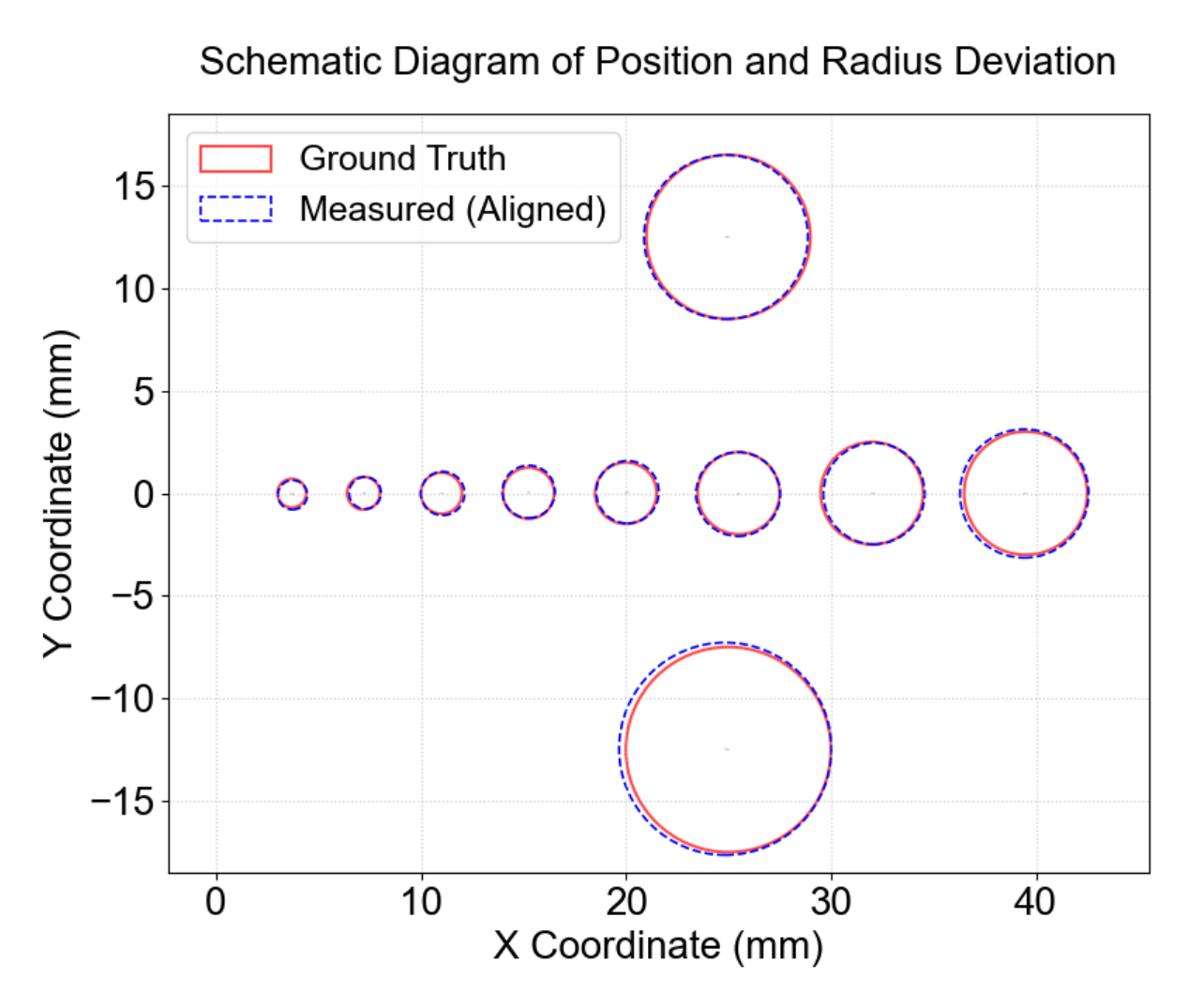}%
 \caption{Experimental results of multi-scale circular geometry estimation. The blue dashed lines represent the estimated circles after Kabsch alignment, while the red solid lines indicate the ground truth from the 3D model.}
 \label{fig11}
\end{figure}

\begin{table}[!t]
	\renewcommand{\arraystretch}{1.3}
	\caption{Quantitative results of position and radius deviation (all units in mm).}
	\centering
	\label{tab3}
	\resizebox{\columnwidth}{!}{
		\begin{tabular}{l c c c c c}
			\hline\hline \\[-3mm]
			\multicolumn{1}{c}{No.} & 1 & 2 & 3 & 4 & 5 \\ \hline
			Pos. Error & 0.0924 & 0.0756 & 0.0625 & 0.0729 & 0.0965 \\
			Rad. Error & 0.0159 & -0.0214 & 0.0663 & 0.0330 & 0.0242 \\ [1.4ex] \hline
			\multicolumn{1}{c}{No.} & 6 & 7 & 8 & 9 & 10 \\ \hline
			Pos. Error & 0.0539 & 0.1068 & 0.0671 & 0.1195 & 0.1572 \\
			Rad. Error & 0.0513 & -0.0280 & 0.1350 & -0.0046 & 0.1799 \\ [1.4ex] \hline
			\textbf{Overall} & \multicolumn{2}{l}{\textbf{Pos. RMSE: 0.0952}} & \multicolumn{3}{l}{\textbf{Rad. Mean: 0.0452}} \\ [1.4ex]
			\hline\hline
		\end{tabular}
	}
\end{table}


The performance is evaluated based on two metrics: position deviation and radius deviation. We employ the Kabsch algorithm \cite{Kabsch} to align the estimated coordinates in the robot frame with the ground truth positions from the 3D model. 

As visualized in Fig. \ref{fig11}, the estimated circular distributions closely align with the design specifications. Quantitative results are summarized in Table \ref{tab3}. The system achieves a RMSE of 0.0952 mm for position estimation and an average radius deviation of 0.0452 mm (MSE). These results demonstrate that SpikingTac can provide high-precision tactile feedback sufficient for micro-scale geometric perception and localization tasks, even when initialized with significant positional uncertainty.

\section{Conclusion}

This article has demonstrated SpikingTac, a neuromorphic visuotactile sensor that bridges the gap between high-speed asynchronous sensing and compact system integration. By synthesizing a global dynamic state map with an unsupervised denoising framework, we have established a robust pipeline capable of maintaining a 1000~Hz perception rate and 350~Hz tracking frequency. The proposed hysteresis-aware incremental update law, featuring a spatial gain damping mechanism, effectively resolves the long-standing trade-off between the high sensitivity of event-driven acquisition and the viscoelastic lag inherent in elastomeric materials.

Our extensive experimental evaluation confirms that the SpikingTac architecture provides superior physical fidelity and temporal consistency. The system exhibits remarkable zero-point stability, characterized by a 100\% return-to-origin success rate and a negligible mean bias of 0.8039 pixels, even when subjected to severe torsional strain. Furthermore, the 5-fold improvement in dynamic response--limiting obstacle-avoidance overshoot to just 6.2~mm--underscores the sensor's advantages over traditional frame-based counterparts in high-velocity interactions. With sub-millimeter geometric precision in micro-feature localization and sizing, SpikingTac offers a high-bandwidth, high-transparency sensing paradigm. This work paves the way for advanced perception in high-speed collaborative robotics and precision industrial automation where low-latency tactile feedback is paramount.


\end{document}